\documentclass[conference]{IEEEtran}

\makeatletter
\newcommand{\newlineauthors}{%
  \end{@IEEEauthorhalign}\hfill\mbox{}\par
  \mbox{}\hfill\begin{@IEEEauthorhalign}
}
\makeatother

\IEEEoverridecommandlockouts
\usepackage{cite}
\usepackage{amsmath,amssymb,amsfonts}
\usepackage{algorithmic}
\usepackage{graphicx}
\usepackage{textcomp}
\usepackage{xcolor}
\usepackage{multicol}

\def\BibTeX{{\rm B\kern-.05em{\sc i\kern-.025em b}\kern-.08em
    T\kern-.1667em\lower.7ex\hbox{E}\kern-.125emX}}
\begin{document}

\title{Co-occurrences using Fasttext embeddings for word similarity tasks in Urdu}

\author{
\IEEEauthorblockN{Usama Khalid}
\IEEEauthorblockA{\textit{Department of Computer Science} \\
\textit{AIM Lab, NUCES (FAST)}\\
Islamabad, Pakistan \\
usama.khalid@nu.edu.pk}
\and
\IEEEauthorblockN{Aizaz Hussain}
\IEEEauthorblockA{\textit{Department of Computer Science} \\
\textit{AIM Lab, NUCES (FAST)}\\
Islamabad, Pakistan \\
aizaz.hussain@nu.edu.pk}
\and
\IEEEauthorblockN{Muhammad Umair Arshad}
\IEEEauthorblockA{\textit{Department of Computer Science} \\
\textit{AIM Lab, NUCES (FAST)}\\
Islamabad, Pakistan \\
umair.arshad@nu.edu.pk}
\newlineauthors
\IEEEauthorblockN{Waseem Shahzad}
\IEEEauthorblockA{\textit{Department of Computer Science} \\
\textit{AIM Lab, NUCES (FAST)}\\
Islamabad, Pakistan \\
waseem.shahzad@nu.edu.pk}
\and
\IEEEauthorblockN{Mirza Omer Beg}
\IEEEauthorblockA{\textit{Department of Computer Science} \\
\textit{AIM Lab, NUCES (FAST)}\\
Islamabad, Pakistan \\
omer.beg@nu.edu.pk}
}

\maketitle

\begin{abstract}
Urdu is a widely spoken language in South Asia. Though immoderate literature exists for the Urdu language still the data isn’t enough to naturally process the language by NLP techniques. Very efficient language models exist for the English language, a high resource language, but Urdu and other under-resourced languages have been neglected for a long time. To create efficient language models for these languages we must have good word embedding models. For Urdu, we can only find word embeddings trained and developed using the skip-gram model. In this paper, we have built a corpus for Urdu by scraping and integrating data from various sources and compiled a vocabulary for Urdu language. We also modify fasttext embeddings and N-Grams models to enable training them on our built corpus.  We have used these trained embeddings for a word similarity task and compared the results with existing techniques. The datasets and code is made freely available on GitHub.\footnote{https://github.com/usamakh20/wordEmbeddingsUrdu}.
\end{abstract}


\begin{figure}[t]
 \includegraphics[width=0.99\linewidth]{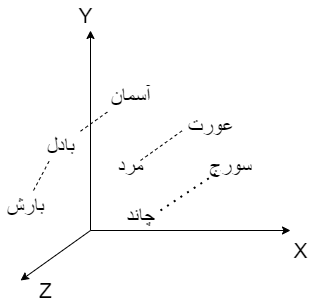}
 \centering
 \caption{N dimensional visualization of word vectors in 3D space.}\label{Fig:image01}
\end{figure}

\section{Introduction}
\begin{figure*}[t!]
 \includegraphics[width=\linewidth]{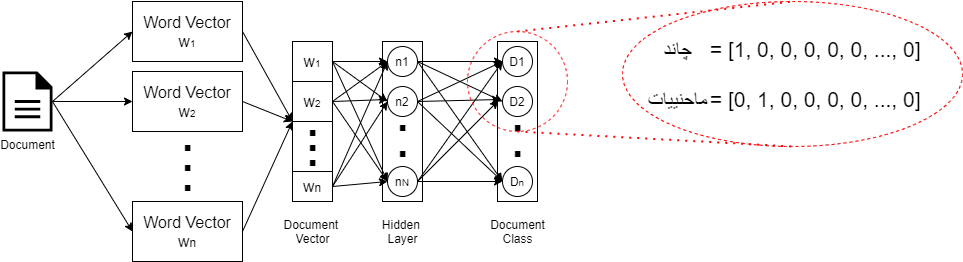}
 \centering
 \caption{An overview of the Fasttext architecture.}\label{Fig:fasttext}
\end{figure*}

Urdu language originated back in 12\textsuperscript{th} with an Indo-Aryan vocabulary \cite{beg2001memory} base and is a mixture of Arabic and Persian. Urdu language is widely spoken and written in the South Asian region with more than 170 million speakers specifically in Pakistan and India. Despite this Urdu \cite{naeem2020deep} is considered a low resourced language because of insufficient data \cite{javed2020collaborative} as compared to English and other widely spoken languages. In recent times the paradigm has been shifted towards the development of efficient models for low resource languages \cite{beg2010graph}.
Many deep learning and machine learning techniques are used to train language models for the derivation of semantics from given textual data \cite{javed2019fairness}. To derive meaningful information from the text it is useful to find out the relation between words. For example, as shown in Fig. \ref{Fig:fasttext}, words are clustered together based on their similarity. Language models store this information which can then be used for many downstream tasks.\\

Machines cannot understand the language \cite{zahid2020roman} in the way we do so data cannot be used as it is passed into the network, instead, each word is converted into a $N$ dimensional vector. These representations are known as word embeddings. An example representation of these embeddings are shown in Fig. \ref{Fig:image01}. The words are projected from an $n$ dimension to 3D \cite{beg2006maxsm,seth2006achieving}. The words with related meanings tend to appear close together. The word embeddings are the baseline for any natural language processing task e.g., transliteration, natural language generation, understanding user inputs, etc \cite{bangash2017methodology}. All these vectors together combined show how much a word is similar to others in a given vector space \cite{beg2019algorithmic}. For Urdu, a lot of work has been done on semantic analysis and sentence classification however there are no studies that show the performance analysis of word embeddings models on the Urdu language \cite{khawaja2018domain}. Unlike the studies conducted for widely spoken languages, in this paper, we use different word embedding models to compute similarity scores for words in the Urdu language.\\

In this paper, we used a freely available Urdu news text corpus COUNTER \cite{sharjeel2017counter,zafar2019constructive}, which contains data from 1200 documents collected from different news agencies of Pakistan. We have then modified existing Fasttext and n-grams approaches to be applied to Urdu data and we train and provide embeddings. Additionally we compare our trained model and embeddings to previously available skip-gram \cite{haider2018urdu,qamarrelationship} technique on a word similarity task \cite{baigawan,qamarrelationship}.\\

This paper is organized in multiple sections which are as follows: In section 2, we will look into the related research. In section 3, we discuss methodologies and the experimental hypothesis. In section 4, we will look at the experimentation results. Finally, in section 5 we will summarize our work and discuss the possible contributions and future directions.

\section{Literature Review}
A lot of work have been done in Urdu language in terms of POS tagging, Sentimental Analysis, NER, Stemmer, MT, Topic Modeling \cite{khan2019part, mukhtar2018urdu, malik2017urdu, gupta2015design, shakeel2018framework,arshad2019corpus,nacem2020subspace,majeed2020emotion,zahid2020roman,khawaja2018domain,rani2015case,baigahmed,beg2001memory} but not much work has been done in word embeddings for Urdu. These words embeddings play a major role in natural language understanding. Multiple language embedding training architectures have been introduced i.e. BERT \cite{devlin2018bert}, Word2Vec etc.\\

There are many ways in which words vectors can be represented among them one is one hot encoding vector representation \cite{hancock2020survey}. In one hot encoding the vectors are represented as long binary vector representations of words \cite{majeed2020emotion}.

% \begin{figure}[!htb]
%  \includegraphics[width=0.75\linewidth]{Images/One Hot Encoding.png}
%  \centering
%  \caption{In this figure the list of numbers against each Urdu word is the word vectors while each dimension is the binary representation of word at the given index in the vector space.}\label{Fig:image05}
% \end{figure}

To formulate one hot vectors for a corpus, they can be aggregated to form the BoW (Bag of Words) representation \cite{zhang2010understanding,beg2006performance}. The bag of words maintain a dictionary of all possible words in the language and keep track of the frequency of the word encountered in the particular corpus.

% \begin{figure}[!htb]
%  \includegraphics[width=1\linewidth]{Images/Bag of words.png}
%  \centering
%  \caption{This figure shows the list of bag of words in multiple documents. There can be N documents and N+1 columns in the table maintaining the bag of words. Whereas, each document column in the table shows the frequency of particular word in the given document.}\label{Fig:image03}
% \end{figure}

The problem with BoW is that it doesn't keep track of words similarity and contextual meaning. So to solve the words similarity problem TF-IDF (Term Frequency - Inverse Document Frequency) \cite{das2018improved,farooq2019bigdata} has been introduced. It associates each word in a document with a number which is a measure of how relevant that word is. Based on this similarity of words, one can compare the similarity of multiple documents together.\\

% \begin{equation}\label{tfidf}
%     TF-IDF = TF(t, d) \times IDF(t)
% \end{equation}

% \begin{equation}\label{equation_tfidf}
%     W\subscript{x,y} = t\subscript{x,y} \times Log(\frac{N}{df\subscript{x}})
% \end{equation}

% The equations \ref{tfidf} and \ref{equation_tfidf}, shows the formulation of TF-TDF. In euation \ref{equation_tfidf} tf\subscript{x, y} shows the frequency of x in y, N shows the number of documnets and df\subscript{x} is the number of documents containing x.  

% \subsection{Word2Vec}
Word2Vec is a fusion of two architectures i.e. CBOW and Skip-Gram \cite{mikolov2013distributed}. These architectures are designed to be mirror images of one another \cite{zafar2020search}. The CBOW the model tries to predict the closest context to the input word while the Skip-Gram model tries to predict the closest words to the input word.

% \begin{figure}[t]
%  \includegraphics[width=1\linewidth]{Images/word2vec.png}
%  \centering
%  \caption{The figure shows the architecture of Word2Vec which consists of two architectures i.e. CBOW \& Skip-Gram. CBOW predicts the word from the provided list of words meanwhile, on given words as input, Skip-Gram predicts the list of words closely related.}\label{Fig:image04}
% \end{figure}

\begin{figure*}[t!]
 \includegraphics[width=1\linewidth]{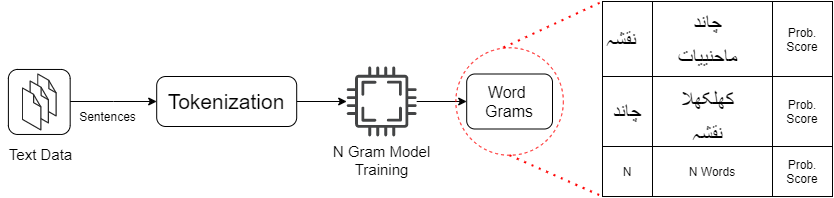}
 \centering
 \caption{An overview of the N-Gram model used in this research.}\label{Fig:ngram}
\end{figure*}

Word embeddings help us considerably improve NLP techniques for low resourced languages. In context of Urdu, the only words embeddings present in literature are that of Skip-Gram\cite{haider2018urdu}. To create a large sample word embeddings for urdu, 140 million sentences in Urdu were used. To check the accuracy of learned embeddings, the closest neighboured words were analyzed w.r.t different words in the vector space \cite{beg2008critical}, context window sizes and their performance on Urdu transliteration.\\

% \subsection{NGrams}
The basic idea behind N-Gram language model is to assign probability to each word in a given sequence of words \cite{pauls2011faster,beg2007flecs}. In word embeddings the words are dissected into multiple N number of chunks and then these chunks are assigned probabilities. Using these probabilities the closest context of a word can be calculated in the vector space. By the calculation of probabilities, this model is very helpful in Natural Language Generations, sentence completion \cite{zafar2019using}, sentence correction etc. The main issue of N-Gram model is that it is sensitive to the training corpus. Many models have been introduced which combine N-Grams and neural networks to overcome the problems of N-Gram and generate more accurate results.

% \subsection{Fasttext}
Fasttext is primarily an architecture developed by facebook for text classification \cite{joulin2016fasttext}. Fasttext works on the principal of Word2Vec and n-grams technique. In word2vec the text is feed into the Neural Network individually. However in Fasttext the words are divided into several sub words and then feed into the Neural Networks. Consider the word apple and we have to dissect this word into tri-grams then the resultant output would be app, ppl, and ple \cite{farooq2019melta}. The word vector for apple will be the sum of all these tri-grams. After training the Neural Network on the training data, we get the word vector for each n-grams and later these n-grams can be used to relate other words. For rare words can be mapped as there will be many overlapping n-grams which appeared in other words.

\section{Methodology}
We used two methods to train our models on word embeddings, Fasttext and N-Grams \cite{alvi2017ensights}. Fig. \ref{Fig:ngram} shows the working of modified N-Gram model used in this research. The n-gram model converts the document into tokens and stores these tokens in a dictionary based on the co-occurrences of words. That is the number of times a token $t_i$ appears next to a token $t_j$ is stored in a co-occurrence dictionary. Against each key there is a are multiple word vectors with the probability score of its occurrence. In fasttext a document is tokenized and passed through a network. The network learns weights which can be extracted as word embeddings. Fig. \ref{Fig:fasttext} shows how words are propagated through the network to extract embeddings for Urdu. In next sections we will discuss in detail about the dataset, experimentation and results.

\subsection{Corpus}\label{AA}
We have used the Urdu Monolingual corpus \cite{jawaid2014tagged} containing 54 million sentences, 90 million tokens and 129K unique tokens. In the preprocessing step we removed all special characters such as brackets, single/double quotes and spaces \cite{thaver2016pulmonary}. All these special characters are replaced by spaces. As a second step consecutive occurring spaces of two or more are matched and replaced by a single space character. 

\subsection{Techniques}
We have used two techniques for t \cite{uzair2019weec}raining, namely ngrams and Fasttext \cite{beg2013constraint}.
The ngram technique requires data to be separated sentence by sentence where each sentence is broken down into a list of words \cite{tariq2019accurate}. After separating into list of words we remove common stop words. Similar prepossessing is applied for Fasttext, however the fasttext python package has the tokenizer and stop word removal tool builtin. The complete architecture is given in figure \ref{Fig:fasttext} and \ref{Fig:ngram}.

\subsection{Hyper Parameters}
The Fasttext technique has four main hyper parameters \cite{zafar2018deceptive} that we can tune. Vector dimension represents the length of the vector size to represent a single word. Larger vectors capture more information \cite{sahar2019towards} but are harder to train and cost more data \cite{koleilat2006watagent}.  Epochs is the number of times the model trains on a batch of data. The larger the corpus the lesser number of times it may have to be iterated. Learning rate is a measure of how quickly the model should converge to a solution \cite{beg2009flecs}. Sub words length specifies the length of substrings \cite{asad2020deepdetect} to consider for different processing tasks like resolving out of vocabulary words.\par

For the current study we have used the default parameters for Fasttext which are 
\begin{itemize}
    \item Vector dimension : 100
    \item Epochs : 5
    \item Learning rate : 0.05
    \item Sub words length : min=3 \& max=6
\end{itemize}

The ngrams technique only has a single hyper-parameter namely the number of consecutive words or grams to train.\par

\section{Results and Discussion}

For the evaluating the similarity of learned word representations We use Urdu translated version of corpora SimLex-999 \cite{hill2015simlex} and WordSim-353 \cite{agirre2009study}. SimLex-999 is a gold standard dataset for evaluating word embeddings. It contains 999 noun, adjective and verb triplets in a concrete \cite{dilawar2018understanding} and abstract form. The dataset is designed to evaluate similarity of words rather than relatedness and contains similarity score for words. The WordSim-353 dataset \cite{javed2020alphalogger} contains relatedness scores for 353 word pairs.  \par

These datasets have been translated to urdu using ijunoon's translation service\footnote{https://translate.ijunoon.com/} and made available. For calculating the similarity and relatedness of words we use the Spearman correlation coefficient \cite{spearman2010proof}. The difference between the predicted score and actual score is \textit{d} and \textit{n} is the number of examples. 

\begin{figure}[t]
 \includegraphics[width=0.99\linewidth]{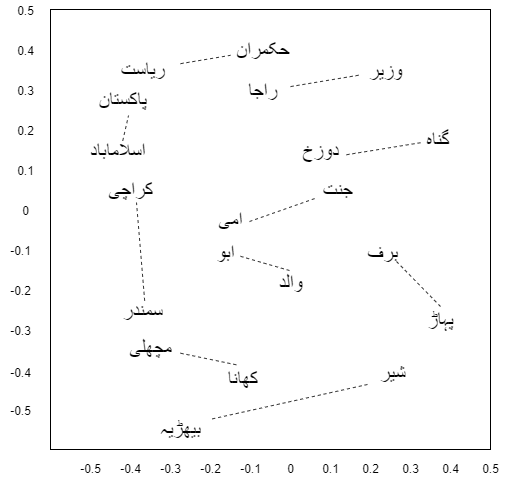}
 \centering
 \caption{In this figure, some word embeddings produced by fasttext are mapped to a 2D space which shows how words are related and how they appear in close proximity to each other.}\label{Fig:image02}
\end{figure}

\begin{equation}
    r\textsubscript{s} = 1 - \frac{6  \Sigma\textsuperscript{n}\textsubscript{i}   d\textsuperscript{2}\textsubscript{i}}{n(n\textsuperscript{2} - 1)}
\end{equation}

\begin{figure}[!htb]
\begin{center}
\begin{tabular}{ |c|c|c|c| } 
  \hline
    & WordSim-353 & SimLex-999 \\
    \hline
    Fasttext & 0.462 & \textbf{0.743} \\ 
    bigrams & 0.188 & 0.156 \\ 
    skip-gram\cite{haider2018urdu}& 0.492 & 0.293 \\ 
    Fasttext English\cite{zhelezniak2019correlation} & 0.84 & 0.76 \\
    \hline
\end{tabular}
\end{center}
\caption{We compare our results with skip-gram\cite{haider2018urdu} embeddings that were evaluated on the same dataset. The Fasttext embeddings are trained for 100 dimensional vectors so the results of skip-gram are also for 100 dimensional vectors for a fair comparison.}\label{table:1}
\end{figure}

The bigrams similarity measure as expected produces a low correlation score, this is also because correlation is only computed for exact word matches from the corpus which are comparatively very less as compared to Fasttext for WordSim-353 and SimLex-999. The Fasttext technique outperforms skip-gram based technique \cite{haider2018urdu} for the SimLex-999 task however slightly under-performs in WordSim-353.

\section{Conclusion}

The advent of Word Embedding techniques \cite{karsten2007axiomatic} was no less than a revolution in the field of NLP. It enabled the representation of words in a digital form (vectors) that computers can understand and perform mathematical calculations on, like the famous example King - Man + Woman = Queen. It also established the ground work for modern Deep attention based models and Transformers in the field of NLP.  \par

Urdu has for long remained an Under resourced language which has caused many proposed state-of-the-art techniques to under perform when being applied to Urdu corpora. It can also be seen in Fig. \ref{table:1} that performance of Fasttext on Urdu corpora is nowhere near to that of English. In this research we have proposed Word co-ocurrences using bigrams and Fasttext word embeddings trained using the COUNTER corpus and have evaluated our approach on WordSim-353 and SimLex-999 similarity tasks and compared that to previously proposed skip-gram technique.\par

In the future work can be done on training these techniques on larger Urdu corpora and evaluate on various tasks like POS Tagging, NER, Machine Translation, sentiment analysis and dependency parsing. In addition to this large corpora have to be proposed for Urdu if we want to at least match the performance of techniques that have been proposed for High resource Languages such as English. We hope that this work will help researchers to produce better techniques in the area of Urdu NLP.

\bibliographystyle{unsrt}
\bibliography{bibliography.bib}

\end{document}